\documentclass[conference]{IEEEtran}
\IEEEoverridecommandlockouts

\usepackage{cite}
\usepackage{amsmath,amssymb,amsfonts}
\usepackage{graphicx}
\usepackage{textcomp}
\usepackage{xcolor}
\usepackage{booktabs}
\usepackage{multirow}
\usepackage{float}
\usepackage{stfloats}
\usepackage{url}
\usepackage[hidelinks]{hyperref}

\usepackage{tikz}
  \newcommand*\circnum[1]{\tikz[baseline=(c.base)]\node[draw,circle,inner sep=1pt](c){#1};}

\begin{document}

\title{MambaGaze: Bidirectional Mamba with Explicit Missing Data Modeling for Cognitive Load Assessment from Eye-Gaze Tracking Data}

\author{
\IEEEauthorblockN{
Amir Mousavi\textsuperscript{1},
Mohammad Sadegh Sirjani\textsuperscript{1},
Erfan Nourbakhsh\textsuperscript{1},
Mimi Xie\textsuperscript{1}\\
Rocky Slavin\textsuperscript{1},
Leslie Neely\textsuperscript{3},
John Davis\textsuperscript{2},
John Quarles\textsuperscript{1}
}

\IEEEauthorblockA{
\textsuperscript{1}Department of Computer Science, College of AI, Cyber and Computing,\\
The University of Texas at San Antonio\\
\{seyedamir.mousavi, mohammadsadegh.sirjani, erfan.nourbakhsh\}@utsa.edu\\
\{mimi.xie, rocky.slavin, john.quarles\}@utsa.edu
}

\IEEEauthorblockA{
\textsuperscript{2}Department of Educational Psychology, College of Education and Human Development,\\
The University of Texas at San Antonio\\
john.davis2@utsa.edu
}

\IEEEauthorblockA{
\textsuperscript{3}Department of Neuroscience, Developmental and Regenerative Biology, College of Sciences,\\
The University of Texas at San Antonio\\
leslie.neely@utsa.edu
}
}

\maketitle

\begin{abstract}
Real-time cognitive load assessment from eye-tracking signals could enable adaptive human-centered AI in safety-critical applications such as driver vigilance monitoring or automated flight deck assistance, yet two challenges persist: handling frequent data missingness from blinks and tracking failures, and efficiently modeling long-range temporal dependencies. We propose MambaGaze (Bi-Mamba), a framework that addresses these challenges through (1)~XMD encoding, which augments raw features with observation masks and time-deltas to explicitly model data uncertainty, and (2)~bidirectional Mamba-2, which captures temporal dependencies with linear computational complexity. Experiments on CLARE and CL-Drive datasets under leave-one-subject-out evaluation show that MambaGaze achieves 77.1\% accuracy and 59.2\% macro-F1 on CLARE, and 69.4\% accuracy and 51.5\% macro-F1 on CL-Drive, attaining the highest average LOSO macro-F1 (55.3\%) across all ten compared models. Input-stream ablation indicates that log-scaled time-deltas are the strongest single channel in our setting, and combining all three XMD streams provides consistent gains of 5--20\,pp macro-F1. Edge deployment benchmarks on three NVIDIA Jetson Orin platforms show real-time inference at 27--36\,FPS with power consumption below 6.6\,W, supporting feasibility for embedded cognitive load monitoring.
\end{abstract}

\begin{IEEEkeywords}
cognitive load, eye-tracking, Mamba, state space models, missing data, edge deployment
\end{IEEEkeywords}

\section{Introduction}

\begin{figure}[t]
    \centering
    \includegraphics[
        width=\columnwidth,
        height=\columnwidth,
        keepaspectratio
    ]{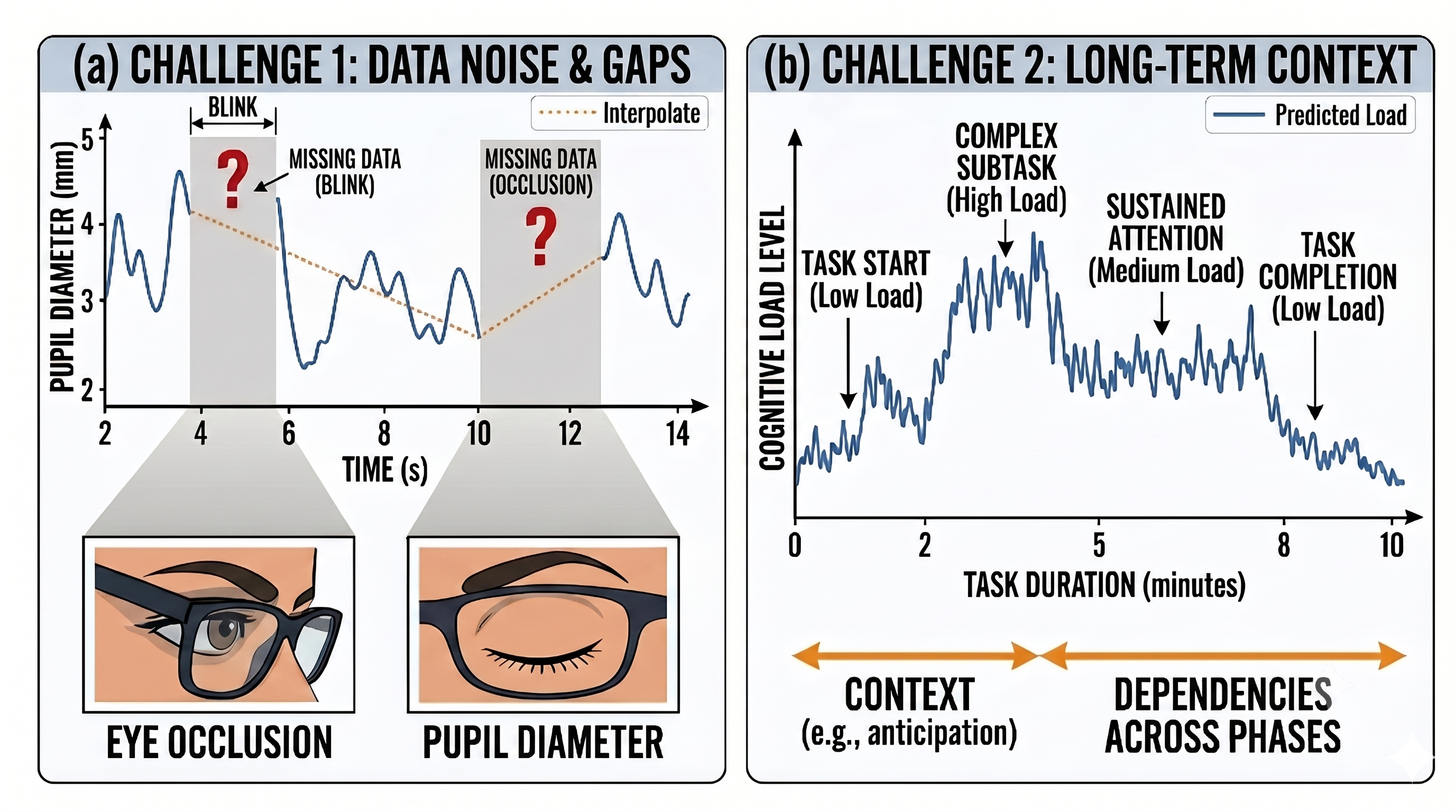}
    \caption{Key Challenges in Real-Time Cognitive Load Assessment from Eye-Tracking Data.}
    \label{fig:intro}
    \vspace{-1em}
\end{figure}

Cognitive load assessment from physiological signals has emerged as a critical research direction with applications in adaptive interfaces, educational technology, and safety-critical systems~\cite{cross_subj_cl_pupil,index_cogload_pupil_resp}. Among various physiological indicators, eye-tracking signals, such as pupil diameter, gaze dynamics, fixations, and saccades, provide rich temporal cues that correlate with cognitive states~\cite{etra2025decoding,scientific2025pupil}. However, effective assessment from such signals via deep learning-based classification requires addressing two key challenges: (1) representing incomplete observations arising from blinks, occlusions, and tracking failures, and (2) capturing long-range temporal dependencies efficiently under strict latency constraints for edge deployment (see Figure~\ref{fig:intro}).

Existing deep learning approaches typically treat missing data as a preprocessing problem, applying interpolation, forward-fill, or deletion before model input~\cite{sarkar2023deep,little2020statistical}. Additionally, window-based aggregation over fixed intervals~\cite{bhatti2024clare} reduces temporal granularity. These strategies distort important fine-grained dynamics for cognitive state inference or discard "informative missingness", patterns where the extended gaps of data itself may indicate specific gaze behaviors or disengagement.

Moreover, conventional approaches lose valuable information about observation reliability and introduce systematic bias, or become heavily reliant on manual feature engineering. While these methods report strong offline validation performance, they have not been validated for edge deployment or real-time processing constraints that practical cognitive monitoring systems require.

For temporal modeling, recurrent architectures suffer from vanishing gradients, while Transformer-based approaches have quadratic complexity that limits scalability~\cite{pi_pp_cl_vr} for high-frequency physiological streams and prevents efficient edge implementation. Recent state space models such as Mamba-2~\cite{dao2024mamba2}, achieve linear complexity while maintaining strong long-range modeling capabilities. However, adapting these models for classification tasks with incomplete data remains unexplored.

In this paper, we propose a novel architecture combining \textbf{Timeseries-Masks-Deltas (XMD) input encoding}~\cite{che2018recurrent} with \textbf{bidirectional Mamba-2} model to enhance cognitive load classification for practical deployment with low-latency requirements. We address the challenge of asynchronous sensor and processing sampling by constructing a tensor that augments resampled feature values (X) with binary masks (M) indicating the arrival of fresh data observation, and time deltas (D) that quantify the temporal gap since the last observed value. This representation is processed by a bidirectional Mamba-2 model, which aggregates temporal context from both forward and reverse directions.  This framework allows the model to distinguish between genuine signal updates and artifacts, maintain the linear complexity of State Space Models (SSMs), and utilize future context to stabilize physiological state inference, even when input data is sparse or irregular.

Our main contributions can be summarized as follows:
\circnum{1} We present an integration of explicit missingness representations (XMD) with selective State Space Models for eye-tracking data. By embedding observation masks and time intervals directly into the input, the model can use missingness patterns rather than treating them only as noise to be removed. \circnum{2} We introduce a bidirectional formulation of Mamba-2 that captures global temporal context, incorporating both past history and retrospective future information without the quadratic cost of Transformers. This targets efficient processing of extended recordings on edge devices. \circnum{3} Experiments show that our approach achieves the highest LOSO macro-F1 (55.3\%) across ten compared models, with 77.1\% accuracy and 59.2\% macro-F1 on CLARE and 69.4\% accuracy on CL-Drive under leave-one-subject-out cross-validation. The model maintains real-time inference latencies of 28--37\,ms across three NVIDIA Jetson Orin platforms.

\section{Related Works}
\label{sec:related}

\subsection{Eye-Tracking Signals for Cognitive Load Assessment}

Eye-tracking measurements provide temporal indicators of cognitive demand across multiple timescales. Relevant biomarkers include pupil diameter dynamics and eye movement behaviors including fixation patterns and saccadic characteristics~\cite{etra2025decoding,scientific2025pupil}. While these signals enable non-invasive cognitive state monitoring, they are vulnerable to environmental and technical artifacts. Specifically, pupil measurements demonstrate high sensitivity to surrounding light variations and calibration errors, whereas gaze sequences frequently contain gaps resulting from physiological events (blinks) and tracking failures (occlusions, signal dropout)~\cite{little2020statistical}. 

Conventional cognitive load pipelines address these challenges through extensive preprocessing and feature aggregation procedures, which necessarily suppress rapid temporal dynamics and systematically exclude missing segments as measurement noise. The present work adopts an alternative approach: direct classification from raw eye-tracking sequences while explicitly preserving observation reliability metadata, thereby enabling the model to leverage missingness patterns as informative signals rather than artifacts requiring elimination.

\subsection{Deep Learning Architectures in Cognitive Load Benchmarks}

Contemporary benchmark datasets have established deep neural networks as standard baselines for cognitive load recognition from physiological and eye-tracking measurements. The CLARE dataset employs convolutional and Transformer-based architectures as representative sequence modeling baselines, and demonstrates competitive classification performance under cross-validation evaluation protocols~\cite{bhatti2024clare}. Similarly, the CL-Drive benchmark evaluates visual recognition backbones including VGG and ResNet architectures for modeling cognitive load in vehicular driving tasks~\cite{angkan2024multimodal}. 

Despite demonstrating effective representation learning capabilities, these investigations share a common limitation in missing data handling. Specifically, existing approaches employ:
1. Complete-case deletion, removing time windows containing gaps
2. Deterministic imputation methods (linear interpolation, forward-fill propagation)
3. Preprocessing pipelines that execute prior to model input~\cite{sarkar2023deep,little2020statistical}

This design blurs observed and imputed values, limiting the model's ability to exploit missingness patterns. In contrast, XMD preserves this distinction and is evaluated under both K-fold and LOSO settings to better reflect subject-level deployment variability.

\subsection{Missingness-Aware Modeling and Efficient Long-Range Sequence Processing}

Parallel research developments have investigated learning frameworks for incomplete temporal sequences. GRU-D and related recurrent architectures incorporate binary masking indicators and temporal decay functions (time-since-last-observation) into hidden state updates which understand differentiation between observed values and model-generated imputations~\cite{che2018recurrent}. However, standard recurrent neural networks exhibit fundamental limitations in long-context modeling and computational parallelization~\cite{goodfellow2016deep}. 

Transformer architectures address these constraints through self-attention mechanisms, though their quadratic computational complexity ($\mathcal{O}(T^2)$) limits scalability for high-frequency data streams~\cite{vaswani2017attention}. Structured state space models (SSMs) provide an alternative framework with linear time complexity. Recent SSM families including S4 and Mamba have demonstrated effective long-range dependency modeling while maintaining computational efficiency~\cite{gu2021efficiently,gu2023mamba}. The Mamba-2 architecture further advances selective state-space parameterization, enabling input-dependent filtering operations at scale~\cite{dao2024mamba2}.

Despite these methodological advances, existing cognitive load classification frameworks have not integrated: (1) explicit missingness representations encompassing observed values, binary indicators, and temporal decay signals with (2) bidirectional structured state space architectures for eye-tracking sequence processing. While GRU-D~\cite{che2018recurrent} and related approaches employ mask-delta formulations with fixed exponential decay, our integration with Mamba-2's selective parameterization enables input-dependent filtering that adapts to the heterogeneous temporal dynamics of eye-tracking signals.

\section{Methodology}
\label{sec:method}

\begin{figure*}[t]
    \centering
    \includegraphics[width=0.9\linewidth]{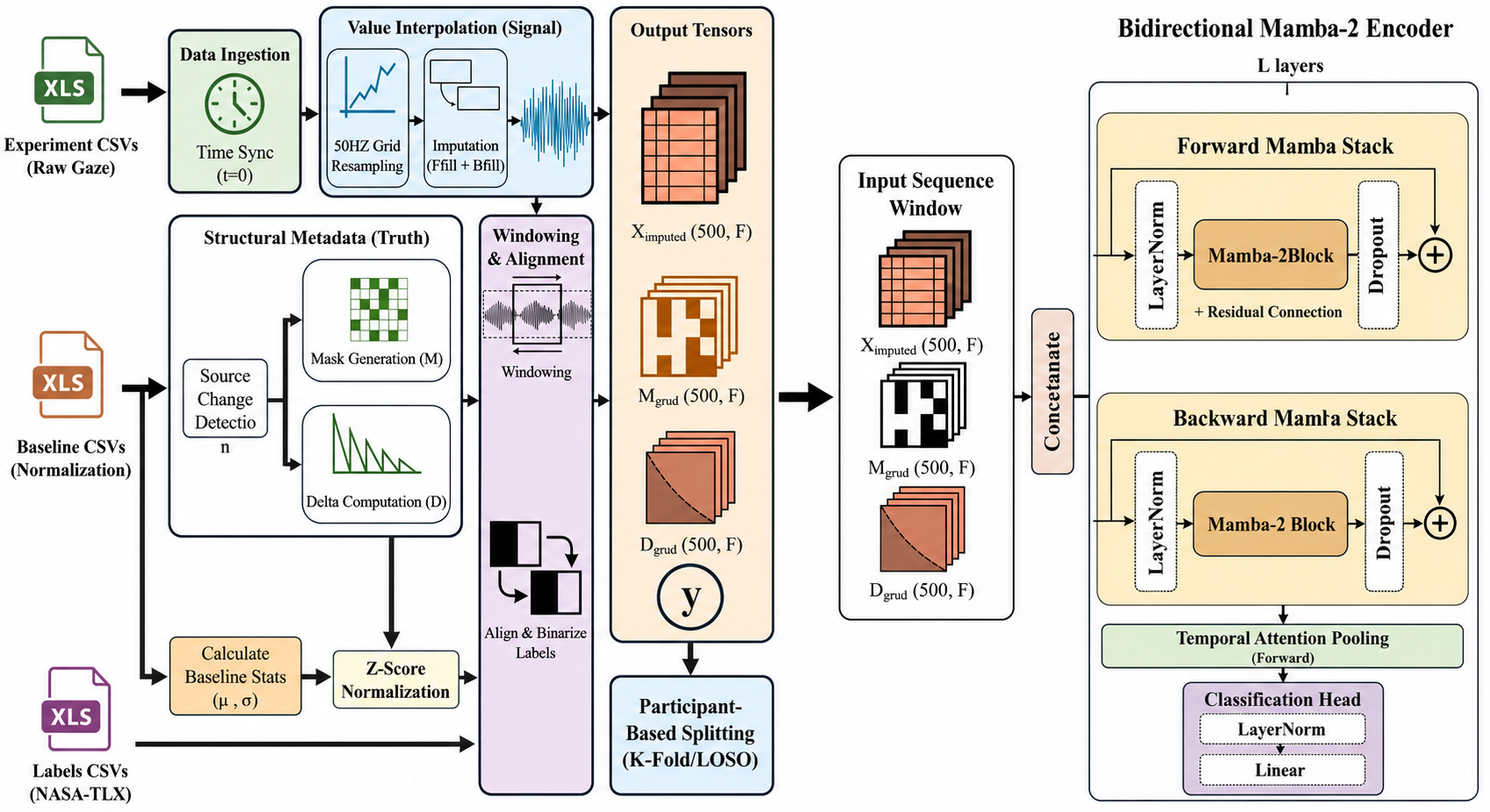}
    \caption{Overview of the MambaGaze pipeline. Raw eye-tracking recordings are first synchronized, cleaned, resampled to a uniform 50\,Hz grid, and normalized using participant-specific baseline statistics. Missing observations are explicitly represented through observation masks ($M$) and time-delta features ($D$), producing XMD-encoded input sequences. The resulting feature, mask, and delta tensors are processed by a bidirectional Mamba-2 backbone that models temporal dependencies in both forward and backward directions. Attention pooling aggregates sequence-level representations, which are passed to a classification head for binary cognitive load prediction.}
    \label{fig:preprocessing}
    \vspace{-1.5em}
\end{figure*}

\subsection{Overview}

This section presents MambaGaze for cognitive load classification from eye-tracking data. The framework processes eye-tracking sequences through four stages:
\begin{itemize}
    \item \textbf{Data Processing Pipeline:} Signal preprocessing, temporal resampling, segmentation, and baseline normalization that preserves observation reliability information (Figure~\ref{fig:preprocessing}, Section~\ref{sec:pipeline});
    \item \textbf{XMD Input Encoding:} Augments raw features with observation masks and time-deltas for explicit missing data modeling (Figure~\ref{fig:preprocessing}, Section~\ref{sec:xmd});
    \item \textbf{Bidirectional Mamba-2 Processing:} Enables efficient long-range temporal modeling with linear complexity using separate forward and backward state space models (Figure~\ref{fig:preprocessing}, Section~\ref{sec:architecture});
    \item \textbf{Attention-based Pooling:} Adaptively aggregates temporal information through learned attention weights for sequence classification (Figure~\ref{fig:preprocessing}, Section~\ref{sec:pooling}).
\end{itemize}

\subsection{Problem Formulation}
\label{sec:problem}

We frame cognitive load assessment as a time-series classification task with inherent missing data challenges. Let $\mathbf{X} \in \mathbb{R}^{F \times T}$ denote eye-tracking signals with $F$ features and length $T$, and $\mathcal{Y} = \{0, 1\}$ represent the binary label space for low or high cognitive load.

\textbf{Missing Data Challenge.} Unlike clinical signals with controlled acquisition, eye-tracking data exhibits frequent missingness due to blinks, gaze aversion, and tracking failures. Standard approaches, deletion or simple imputation, discard valuable reliability information. The missing pattern itself carries signal: frequent blinks may indicate fatigue, while tracking losses correlate with rapid eye movements during high cognitive load.

\textbf{Solution Strategy.} Our goal is to learn a mapping $\Phi: \mathbb{R}^{F \times T} \rightarrow \mathcal{Y}$ that explicitly models observation uncertainty. We address this through: (1) a data processing pipeline that preserves missingness information rather than discarding it, (2) XMD encoding that augments features with observation masks and time-deltas, and (3) a bidirectional state space model that efficiently captures long-range temporal dependencies.

\subsubsection{Data Processing Pipeline}
\label{sec:pipeline}

The data processing pipeline transforms raw asynchronous eye-tracking recordings into fixed-length, normalized windows suitable for sequence modeling while preserving observation reliability information.

\textbf{Signal Preprocessing.} Raw eye-tracking data arrives asynchronously, with multiple sensor channels updating at different times. We extract $F=10$ canonical features: pupil diameter (left/right), gaze coordinates $(x, y)$, velocity, acceleration, fixation indicator, saccade indicator, blink events, and distance. Duplicate timestamps are coalesced per-column using the last valid observation, preventing incorrect merging of asynchronous channel updates.

\textbf{Temporal Resampling.} We resample all signals to a uniform grid at $f_s = 50$ Hz using forward-fill interpolation, yielding $T = 500$ timesteps per 10-second window. Critically, we compute observation masks from the \emph{original} source timestamps via per-channel change detection before resampling, this preserves the true observation pattern rather than the interpolated values.

\textbf{Temporal Segmentation and Label Assignment.} We segment continuous recordings into fixed-length windows $\mathbf{x}_k \in \mathbb{R}^{F \times T}$ of duration $w = 10$ seconds. Cognitive load labels are provided at $\Delta = 10$-second intervals (NASA-TLX scores 1--9, binarized at threshold 5) based on source datasets protocol. For each window $W_k = [t_k, t_k + w)$, we assign the label corresponding to the interval covering the window's midpoint:
\begin{equation}
y(W_k) = y_{i^\star}, \quad i^\star = \left\lfloor \frac{t_k + w/2}{\Delta} \right\rfloor
\end{equation}
This midpoint rule ensures unambiguous label assignments at interval boundaries.

\textbf{Baseline Normalization.} Within each participant, we compute baseline statistics $(\mu_f^{\text{base}}, \sigma_f^{\text{base}})$ from a dedicated resting-state recording. All features are z-score normalized as $\tilde{x}_{t,f} = (x_{t,f} - \mu_f^{\text{base}}) / (\sigma_f^{\text{base}} + \epsilon)$, accounting for inter-individual variability in pupil size, gaze patterns, and sensor calibration.

\subsubsection{XMD Input Encoding}
\label{sec:xmd}

To explicitly model observation uncertainty, we adopt Values-Masks-Deltas (XMD) encoding~\cite{che2018recurrent} that represents missing data within the model input, enabling the network to learn different behaviors for observed versus imputed values.

\textbf{Values.} For each feature $f$, missing values are imputed using forward-fill with baseline initialization:
\begin{equation}
\tilde{x}_{t,f} = \begin{cases}
x_{t,f} & \text{if } m_{t,f} = 1 \\
\tilde{x}_{t-1,f} & \text{if } m_{t,f} = 0 \text{ and } t > 1 \\
\mu_f^{\text{base}} & \text{if } m_{t,f} = 0 \text{ and } t = 1
\end{cases}
\end{equation}

\textbf{Masks.} Binary observation masks $\mathbf{M} \in \{0,1\}^{T \times F}$ indicate data reliability. The mask is computed via change detection on the original (pre-resampling) data: $m_{t,f} = 1$ when feature $f$ receives a genuinely new observation at time $t$, distinguishing true measurements from interpolated values.

\textbf{Deltas.} Time deltas $\mathbf{D} \in \mathbb{R}_{\geq 0}^{T \times F}$ encode elapsed time since the last valid observation:
\begin{equation}
\delta_{t,f} = \begin{cases}
0 & \text{if } m_{t,f} = 1 \\
\delta_{t-1,f} + \Delta t & \text{if } m_{t,f} = 0
\end{cases}
\end{equation}
where $\Delta t = 1/f_s$. We apply log-scaling $\bar{\delta}_{t,f} = \log(1 + \delta_{t,f})$ to compress the dynamic range.

The complete XMD representation concatenates these components:
$\mathbf{Z}_t = [\tilde{\mathbf{x}}_t \| \mathbf{m}_t \| \bar{\boldsymbol{\delta}}_t] \in \mathbb{R}^{3F}$.
This tripling of input dimensionality ($F' = 3F = 30$) provides explicit uncertainty quantification without architectural modifications.

\subsection{Architecture}
\label{sec:architecture}

We employ Mamba-2~\cite{dao2024mamba2}, a selective state space model achieving linear complexity $\mathcal{O}(T)$ compared to the quadratic complexity $\mathcal{O}(T^2)$ of Transformer-based attention. For eye-tracking sequences spanning hundreds of timesteps ($T = 500$ at 50 Hz), this efficiency is essential.

Mamba-2 introduces \emph{selective} state spaces where transition parameters depend on the input content, computing input-dependent parameters $\mathbf{B}_t$, $\mathbf{C}_t$, and step size $\Delta_t$ from each input $\mathbf{x}_t$. This selectivity enables content-aware filtering, allowing the model to learn to retain or forget information based on input relevance. For eye-tracking data, transient events like saccades and blinks require different temporal dynamics than sustained states like fixations.

\subsubsection{Bidirectional Mamba-2 Processing}

Cognitive load assessment benefits from bidirectional context, as pupil dilation may reflect anticipation of upcoming cognitive demands or reaction to recent mental effort. Standard Mamba processes sequences causally (left-to-right), limiting access to future context. We extend Mamba-2 to bidirectional processing using two separate branches with independent parameters.

\textbf{Forward Mamba.} Processes the sequence in causal order as $\mathbf{H}^{\text{fwd}} = \text{Mamba}(\mathbf{Z}_{1:T}; \boldsymbol{\theta}^{\text{fwd}})$.

\textbf{Backward Mamba.} Processes the time-reversed sequence as $\mathbf{H}^{\text{bwd}} = \text{flip}(\text{Mamba}(\text{flip}(\mathbf{Z}_{1:T}); \boldsymbol{\theta}^{\text{bwd}}))$, where $\text{flip}(\cdot)$ reverses the temporal dimension.

Each branch consists of $L$ stacked Mamba-2 layers with pre-normalization and residual connections: $\mathbf{h}_l = \mathbf{h}_{l-1} + \text{Dropout}(\text{Mamba}(\text{LayerNorm}(\mathbf{h}_{l-1})))$. An input projection layer maps the XMD features to the model dimension: $\mathbf{h}_0 = \mathbf{W}_{\text{in}} \mathbf{Z} + \mathbf{b}_{\text{in}}$, where $\mathbf{W}_{\text{in}} \in \mathbb{R}^{D \times 3F}$.

\begin{table*}[t]
  \centering
  \caption{Comparison on CLARE and CL-Drive under LOSO and 10-fold cross-validation.
  Values are mean\,$\pm$\,std (\%) over folds;
  F1$_{\text{M}}$ is macro-F1; Params is the trainable parameter count in thousands~(K).
  Avg averages the two datasets within each protocol;
  rows ordered by LOSO average macro-F1.}
  \label{tab:main_results}
  \renewcommand{\arraystretch}{1.2}%
  \setlength{\tabcolsep}{3pt}%
  \resizebox{\textwidth}{!}{%
  \begin{tabular}{lccccccccccccccccccc}
    \toprule
    & & \multicolumn{6}{c}{\textbf{CLARE}} & \multicolumn{6}{c}{\textbf{CL-Drive}}
    & \multicolumn{6}{c}{\textbf{Avg}} \\
    \cmidrule(lr){3-8} \cmidrule(lr){9-14} \cmidrule(lr){15-20}
    & & \multicolumn{3}{c}{LOSO} & \multicolumn{3}{c}{10-fold}
    & \multicolumn{3}{c}{LOSO} & \multicolumn{3}{c}{10-fold}
    & \multicolumn{3}{c}{LOSO} & \multicolumn{3}{c}{10-fold} \\
    \cmidrule(lr){3-5} \cmidrule(lr){6-8} \cmidrule(lr){9-11} \cmidrule(lr){12-14}
    \cmidrule(lr){15-17} \cmidrule(lr){18-20}
    Model & Params (K) & Acc & F1$_{\text{M}}$ & AUC & Acc & F1$_{\text{M}}$ & AUC
          & Acc & F1$_{\text{M}}$ & AUC & Acc & F1$_{\text{M}}$ & AUC
          & Acc & F1$_{\text{M}}$ & AUC & Acc & F1$_{\text{M}}$ & AUC \\
    \midrule
    CNN & 5679.5
      & 73.0{\scriptsize\,$\pm$16.0} & 47.4{\scriptsize\,$\pm$5.7}  & 53.1{\scriptsize\,$\pm$13.1}
      & 74.3{\scriptsize\,$\pm$9.2}  & 46.9{\scriptsize\,$\pm$4.2}  & 57.5{\scriptsize\,$\pm$6.5}
      & 64.3{\scriptsize\,$\pm$12.4} & 54.6{\scriptsize\,$\pm$9.3}  & 60.3{\scriptsize\,$\pm$12.4}
      & 62.6{\scriptsize\,$\pm$9.3}  & 52.3{\scriptsize\,$\pm$8.1}  & 57.0{\scriptsize\,$\pm$10.2}
      & 68.6 & 51.0 & 56.7 & 68.4 & 49.6 & 57.2 \\
    Transformer & 614.0
      & 74.0{\scriptsize\,$\pm$13.7} & 49.3{\scriptsize\,$\pm$5.0}  & 53.9{\scriptsize\,$\pm$12.9}
      & 70.5{\scriptsize\,$\pm$9.5}  & 47.7{\scriptsize\,$\pm$5.2}  & 54.7{\scriptsize\,$\pm$3.4}
      & 61.0{\scriptsize\,$\pm$10.4} & 53.2{\scriptsize\,$\pm$7.5}  & 56.7{\scriptsize\,$\pm$12.7}
      & 60.1{\scriptsize\,$\pm$6.0}  & 52.7{\scriptsize\,$\pm$7.2}  & 56.1{\scriptsize\,$\pm$8.2}
      & 67.5 & 51.2 & 55.3 & 65.3 & 50.2 & 55.4 \\
    ResNet & 1036.9
      & 72.8{\scriptsize\,$\pm$13.6} & 50.1{\scriptsize\,$\pm$6.0}  & 55.1{\scriptsize\,$\pm$14.3}
      & 69.9{\scriptsize\,$\pm$7.7}  & 50.9{\scriptsize\,$\pm$6.6}  & 58.2{\scriptsize\,$\pm$8.0}
      & 61.7{\scriptsize\,$\pm$10.4} & 52.6{\scriptsize\,$\pm$10.4} & 59.1{\scriptsize\,$\pm$13.2}
      & 58.0{\scriptsize\,$\pm$8.8}  & 50.4{\scriptsize\,$\pm$8.7}  & 51.8{\scriptsize\,$\pm$10.2}
      & 67.2 & 51.3 & 57.1 & 63.9 & 50.7 & 55.0 \\
    VGG & 418.6
      & 72.3{\scriptsize\,$\pm$12.9} & 49.8{\scriptsize\,$\pm$7.0}  & 53.6{\scriptsize\,$\pm$14.1}
      & 66.4{\scriptsize\,$\pm$9.2}  & 52.5{\scriptsize\,$\pm$3.2}  & 57.1{\scriptsize\,$\pm$5.0}
      & 62.0{\scriptsize\,$\pm$12.1} & 53.7{\scriptsize\,$\pm$10.0} & 59.2{\scriptsize\,$\pm$13.2}
      & 58.1{\scriptsize\,$\pm$10.9} & 51.2{\scriptsize\,$\pm$7.5}  & 54.2{\scriptsize\,$\pm$9.1}
      & 67.1 & 51.7 & 56.4 & 62.2 & 51.9 & 55.7 \\
    Uni-S4 & 221.2
      & 76.7{\scriptsize\,$\pm$12.2} & 56.4{\scriptsize\,$\pm$11.3} & 61.8{\scriptsize\,$\pm$17.8}
      & 77.0{\scriptsize\,$\pm$7.0}  & 52.5{\scriptsize\,$\pm$6.7}  & 59.6{\scriptsize\,$\pm$6.5}
      & 70.4{\scriptsize\,$\pm$10.3} & 49.3{\scriptsize\,$\pm$7.9}  & 56.9{\scriptsize\,$\pm$10.6}
      & 69.9{\scriptsize\,$\pm$7.0}  & 46.0{\scriptsize\,$\pm$5.5}  & 54.8{\scriptsize\,$\pm$7.7}
      & 73.5 & 52.8 & 59.3 & 73.4 & 49.2 & 57.2 \\
    Uni-Mamba & 440.5
      & 77.5{\scriptsize\,$\pm$11.5} & 58.3{\scriptsize\,$\pm$8.9}  & 60.3{\scriptsize\,$\pm$17.0}
      & 77.8{\scriptsize\,$\pm$7.0}  & 52.5{\scriptsize\,$\pm$10.5} & 59.2{\scriptsize\,$\pm$7.9}
      & 71.6{\scriptsize\,$\pm$9.6}  & 47.0{\scriptsize\,$\pm$7.4}  & 56.3{\scriptsize\,$\pm$13.0}
      & 65.7{\scriptsize\,$\pm$8.8}  & 48.7{\scriptsize\,$\pm$4.7}  & 52.9{\scriptsize\,$\pm$7.3}
      & 74.6 & 52.7 & 58.3 & 71.7 & 50.6 & 56.0 \\
    Uni-S4D & 220.7
      & 77.3{\scriptsize\,$\pm$12.2} & 56.4{\scriptsize\,$\pm$11.4} & 61.3{\scriptsize\,$\pm$18.0}
      & 77.6{\scriptsize\,$\pm$6.6}  & 54.1{\scriptsize\,$\pm$8.2}  & 59.7{\scriptsize\,$\pm$5.3}
      & 72.5{\scriptsize\,$\pm$9.1}  & 49.9{\scriptsize\,$\pm$6.8}  & 56.1{\scriptsize\,$\pm$13.3}
      & 67.1{\scriptsize\,$\pm$7.6}  & 48.4{\scriptsize\,$\pm$5.8}  & 56.4{\scriptsize\,$\pm$8.0}
      & 74.9 & 53.1 & 58.7 & 72.4 & 51.3 & 58.0 \\
    Bi-S4D & 441.3
      & 76.7{\scriptsize\,$\pm$12.3} & 56.4{\scriptsize\,$\pm$11.3} & 61.3{\scriptsize\,$\pm$17.8}
      & 77.3{\scriptsize\,$\pm$6.7}  & 53.9{\scriptsize\,$\pm$8.0}  & 61.1{\scriptsize\,$\pm$6.9}
      & 71.2{\scriptsize\,$\pm$10.5} & 49.6{\scriptsize\,$\pm$7.5}  & 56.9{\scriptsize\,$\pm$11.4}
      & 66.4{\scriptsize\,$\pm$9.0}  & 49.0{\scriptsize\,$\pm$5.5}  & 53.7{\scriptsize\,$\pm$5.8}
      & 73.9 & 53.0 & 59.1 & 71.9 & 51.4 & 57.4 \\
    Bi-S4 & 442.4
      & 77.2{\scriptsize\,$\pm$12.0} & 56.7{\scriptsize\,$\pm$11.4} & 61.9{\scriptsize\,$\pm$18.2}
      & 77.7{\scriptsize\,$\pm$6.7}  & 53.0{\scriptsize\,$\pm$9.2}  & 60.2{\scriptsize\,$\pm$7.8}
      & 71.7{\scriptsize\,$\pm$9.7}  & 51.9{\scriptsize\,$\pm$8.8}  & 58.0{\scriptsize\,$\pm$12.9}
      & 66.1{\scriptsize\,$\pm$9.0}  & 49.5{\scriptsize\,$\pm$5.2}  & 54.1{\scriptsize\,$\pm$5.5}
      & 74.4 & 54.3 & 60.0 & 71.9 & 51.2 & 57.2 \\
    \textbf{Bi-Mamba (Ours)} & \textbf{881.0}
      & \textbf{77.1{\scriptsize\,$\pm$12.0}} & \textbf{59.2{\scriptsize\,$\pm$9.4}}  & \textbf{62.0{\scriptsize\,$\pm$18.2}}
      & \textbf{77.2{\scriptsize\,$\pm$6.4}}  & \textbf{53.1{\scriptsize\,$\pm$7.6}}  & \textbf{60.0{\scriptsize\,$\pm$6.6}}
      & \textbf{69.4{\scriptsize\,$\pm$11.2}} & \textbf{51.5{\scriptsize\,$\pm$8.6}}  & \textbf{57.9{\scriptsize\,$\pm$11.6}}
      & \textbf{65.4{\scriptsize\,$\pm$6.3}}  & \textbf{50.9{\scriptsize\,$\pm$5.8}}  & \textbf{55.8{\scriptsize\,$\pm$6.6}}
      & \textbf{73.2} & \textbf{55.3} & \textbf{59.9} & \textbf{71.3} & \textbf{52.0} & \textbf{57.9} \\
    \bottomrule
  \end{tabular}%
  }
  \vspace{-1.3em}
\end{table*}

Unlike bidirectional RNNs that concatenate hidden states at each timestep, we keep the forward and backward branches parametrically separate, so one branch encodes how earlier gaze behavior leads into the current state and the other how it resolves into later behavior; the two are combined only at pooling. Because the backward pass needs the full window, each prediction is computed once per completed 10-second segment (dataset protocol), so the window delay is set by the prediction target, not the architecture.

\subsubsection{End-to-End Computation}

For each XMD-encoded window $\mathbf{Z}_k \in \mathbb{R}^{T \times 3F}$, the graph below encodes both directions, pools each sequence, concatenates the contexts, and predicts a probability:
\begin{align}
\mathbf{H}_k^{\text{fwd}} &= \text{Mamba}(\mathbf{Z}_k; \boldsymbol{\theta}^{\text{fwd}}) \label{eq:fwd}\\
\mathbf{H}_k^{\text{bwd}} &= \text{flip}(\text{Mamba}(\text{flip}(\mathbf{Z}_k); \boldsymbol{\theta}^{\text{bwd}})) \label{eq:bwd}\\
\mathbf{c}_k^{\text{fwd}} &= \text{AttnPool}(\mathbf{H}_k^{\text{fwd}}) \in \mathbb{R}^{D} \label{eq:pool_fwd}\\
\mathbf{c}_k^{\text{bwd}} &= \text{AttnPool}(\mathbf{H}_k^{\text{bwd}}) \in \mathbb{R}^{D} \label{eq:pool_bwd}\\
\mathbf{c}_k &= [\mathbf{c}_k^{\text{fwd}} \| \mathbf{c}_k^{\text{bwd}}] \in \mathbb{R}^{2D} \label{eq:concat}\\
\hat{y}_k &= \sigma(g_\psi(\mathbf{c}_k)) \label{eq:pred}
\end{align}
Here $\text{flip}$ reverses time, $D$ is the hidden width, $\text{AttnPool}(\cdot)$ is defined in Section~\ref{sec:pooling}, $\sigma$ is sigmoid, and $g_\psi$ is the classification head.

\subsection{Attention Pooling and Classification}
\label{sec:pooling}

Sequence classification requires aggregating temporal representations into a fixed-size vector. Rather than using the final hidden state (which biases toward recent information) or simple averaging (which weights all timesteps equally), we employ learned attention pooling that adaptively weights timesteps based on their relevance to the classification task.

\subsubsection{Attention Pooling}

For each direction, we compute attention scores using an additive attention mechanism with a learned query:
\begin{align}
e_t &= \mathbf{w}^\top \tanh(\mathbf{W}_a \mathbf{h}_t + \mathbf{b}_a) \\
\alpha_t &= \text{softmax}_t(e_t) = \frac{\exp(e_t)}{\sum_{t'=1}^{T} \exp(e_{t'})} \\
\mathbf{c}^{\text{dir}} &= \sum_{t=1}^{T} \alpha_t \mathbf{h}_t^{\text{dir}}
\end{align}
where $\mathbf{W}_a \in \mathbb{R}^{D \times D}$, $\mathbf{b}_a \in \mathbb{R}^{D}$, and $\mathbf{w} \in \mathbb{R}^{D}$ are learnable parameters. Separate attention modules are used for forward and backward directions, yielding context vectors $\mathbf{c}^{\text{fwd}}$ and $\mathbf{c}^{\text{bwd}}$ that are concatenated to form the final sequence representation $\mathbf{c} \in \mathbb{R}^{2D}$.

The attention weights $\alpha_t$ provide a diagnostic view of which temporal regions receive higher weight. 

\subsubsection{Classification Head Architecture}

The classification head $g_\psi$ is a lightweight module that maps the pooled representation to a binary prediction: $\hat{y} = \sigma(\mathbf{w}_c^\top \text{LayerNorm}(\mathbf{c}) + b_c)$, where $\sigma(\cdot)$ is the sigmoid function, $\mathbf{w}_c \in \mathbb{R}^{2D}$, and $b_c \in \mathbb{R}$ are learnable parameters. Layer normalization stabilizes activations before the final projection.

\subsection{Training Objective}
\label{sec:training}

\subsubsection{Loss Function with Automatic Class Rebalancing}

Let $y_k \in \{0, 1\}$ and $\hat{y}_k \in [0,1]$ denote the window label and predicted probability. To address class imbalance in cognitive load distributions, we minimize a weighted binary cross-entropy loss:
\begin{equation}
\mathcal{L} = -\frac{1}{N}\sum_{k=1}^{N} \left[ w_+ y_k \log(\hat{y}_k) + (1-y_k) \log(1-\hat{y}_k) \right]
\label{eq:bce}
\end{equation}

\textbf{Automatic Class Weighting.} The BCE positive-class multiplier is computed per fold as $w_+ = n_- / n_+$, balancing high-load and low-load terms from the training counts. We monitor macro-F1 on the validation set to detect majority-class collapse and for early stopping.

Training uses AdamW~\cite{LoshchilovH19} ($\text{lr}=10^{-4}$, weight decay$=0$, batch size$=128$, gradient clipping$=0.5$) with mixed-precision (AMP) training. A 5\% validation split monitors macro-F1 for early stopping with patience 15 epochs (max 100 epochs).

\section{Experiments}
\label{sec:experiments}

\subsection{Experimental Setup}

\paragraph{Datasets.} We evaluate on two public cognitive load datasets: CLARE (20 subjects performing cognitive tasks)~\cite{bhatti2024clare} and CL-Drive (15 subjects in simulated driving)~\cite{angkan2024multimodal}. Both provide eye-tracking data with self-reported cognitive load labels at 10-second intervals (1--9 scale), binarized into low (1--4) versus high (5--9) classes.

\paragraph{Preprocessing.} Raw eye-tracking data is resampled to 50\,Hz and segmented into non-overlapping windows of $T=500$ timesteps (10 seconds). We extract $F=10$ features including pupil diameter, gaze position, velocity, acceleration, fixation/saccade/blink indicators, and screen distance. With XMD encoding, each feature is augmented with observation mask and time-delta channels, yielding $F'=30$ input dimensions.

\paragraph{Evaluation Protocol.} We use two cross-validation strategies: (i) Leave-One-Subject-Out (LOSO) to assess generalization to unseen individuals, and (ii) participant-based 10-fold cross-validation. To address class imbalance, we apply dynamic positive class weighting~(W), decision-threshold calibration~(T), and post-hoc prediction flip~(P).

\paragraph{Baselines.} We compare against four architecture families. \emph{Dataset-specific baselines}: CNN and Transformer from CLARE~\cite{bhatti2024clare}, and ResNet and VGG from CL-Drive~\cite{angkan2024multimodal}. \emph{SSM architectural variants}: unidirectional and bidirectional variants of S4~\cite{gu2021efficiently}, S4D~\cite{gu2022parameterization}, and Mamba-2~\cite{dao2024mamba2}, all trained with the identical XMD preprocessing and optimization pipeline, isolating architecture from input representation.

\subsection{Main Results}

Table~\ref{tab:main_results} compares all ten models on both datasets under both protocols.

Bi-Mamba (MambaGaze) achieves the highest average LOSO macro-F1 across both datasets
(55.3\%), ranking first among the ten models.
On CLARE under LOSO, Bi-Mamba reaches 59.2\% F1, on par with
the next-best SSM variant (Uni-Mamba, 58.3\%) and
\mbox{+9.9\,pp} above the Transformer baseline (49.3\%).
On CL-Drive under LOSO, Bi-Mamba achieves the highest accuracy (69.4\%),
though its F1 (51.5\%) is slightly below Bi-S4 (51.9\%)
and VGG (53.7\%), a consequence of the dataset's greater class imbalance
and lower cross-subject consistency (higher LOSO variance).
Across both datasets and both protocols, Bi-Mamba ranks first
in LOSO average F1 (55.3\%) and is within 0.1\,pp of the best 10-fold F1 (52.0\% vs.\ 51.9\% for VGG),
indicating that bidirectional Mamba-2 combined with XMD encoding generalizes reliably
to unseen subjects.
Within the SSM family, bidirectionality improves Mamba and S4 but not S4D,
so we treat directionality as architecture-dependent rather than uniformly beneficial.

\subsection{Ablation Study}

\subsubsection{Training Optimization}

\begin{figure}[t]
    \centering
    \includegraphics[width=\columnwidth]{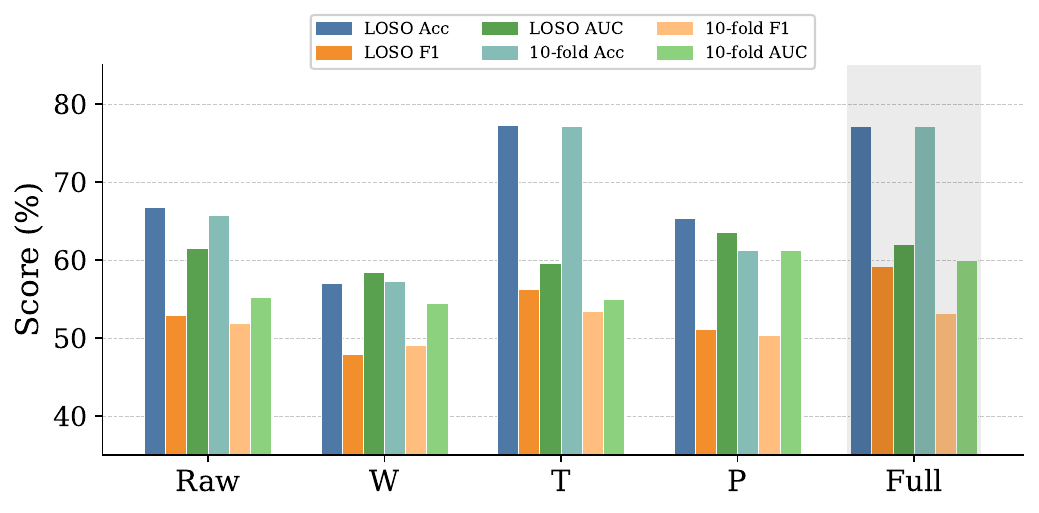}
    \caption{Training optimization ablation for Bi-Mamba on CLARE.
    Each group of bars shows Acc, F1$_{\text{M}}$, and AUC under LOSO and 10-fold CV.
    The shaded group (Full) combines all three techniques.}
    \label{fig:ablation_opt}
\end{figure}

Figure~\ref{fig:ablation_opt} shows the contribution of each optimization technique
on CLARE under LOSO and 10-fold cross-validation.
W denotes positive class weighting in the BCE loss,
T denotes decision-threshold calibration on the validation fold,
and P denotes the post-hoc prediction flip applied when the fold's validation AUC falls below 0.5.

Threshold calibration~(T) provides the largest single-technique gain:
\mbox{+10.6\,pp} accuracy on CLARE-LOSO (66.7\% $\to$ 77.3\%) and
\mbox{+13.2\,pp} on CL-Drive-LOSO (54.7\% $\to$ 67.9\%).
This indicates that the default 0.5 decision boundary is poorly suited to
the imbalanced cognitive load distributions in both datasets.
Class weighting~(W) alone is counterproductive, accuracy \emph{drops}
to 57.0\% on CLARE-LOSO versus 66.7\% raw, suggesting that count reweighting
is poorly matched to the default 0.5 threshold.
Post-hoc flip~(P) primarily improves AUC (63.6\% on CLARE-LOSO) with minimal
effect on accuracy or F1.
Combining all three (Bi-Mamba Full) yields the best F1 on every condition,
suggesting that calibrated thresholds mitigate the instability of W alone.

\subsubsection{Input Stream Ablation}

\begin{figure}[t]
    \centering
    \includegraphics[width=\columnwidth]{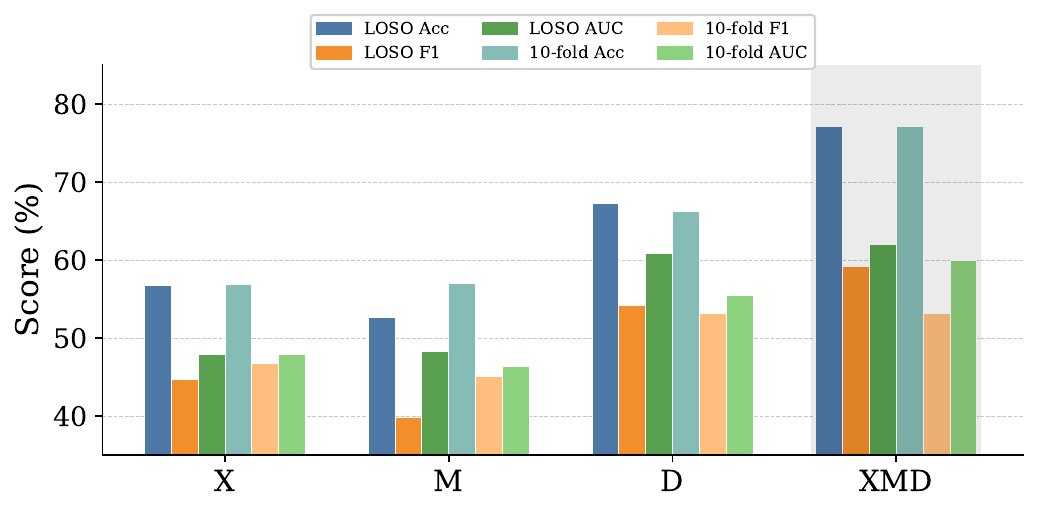}
    \caption{Input stream ablation for Bi-Mamba on CLARE.
    Each group of bars shows Acc, F1$_{\text{M}}$, and AUC under LOSO and 10-fold CV
    for each single-stream variant (X, M, D) and the full XMD combination.}
    \label{fig:input_ablation}
\end{figure}

Figure~\ref{fig:input_ablation} shows the contribution of each XMD input channel
on CLARE: imputed values~(X), observation masks~(M), and log-scaled time-deltas~(D).

Time-deltas~(D) are the strongest single channel in this ablation, achieving 54.2\% F1 on
CLARE-LOSO, only 5\,pp below the full XMD model.
This suggests that the temporal spacing of observations is a strong signal for
cognitive state, capturing gaze irregularities that raw interpolated values
(X alone: 44.7\% F1) cannot recover.
The binary mask~(M) alone is weakest for CLARE (39.8\% F1), but surprisingly
strong for CL-Drive AUC (66.7\%), suggesting that the presence/absence pattern
carries dataset-specific information about driving-induced gaze lapses.
Combining all three streams (XMD) yields consistent gains of
\mbox{+5\,--\,20\,pp} F1 over every single-stream baseline, indicating that
imputed values, missingness patterns, and temporal gaps provide complementary signals.

\subsection{Edge Deployment}

\begin{figure*}[t]
  \centering
  \includegraphics[width=\linewidth]{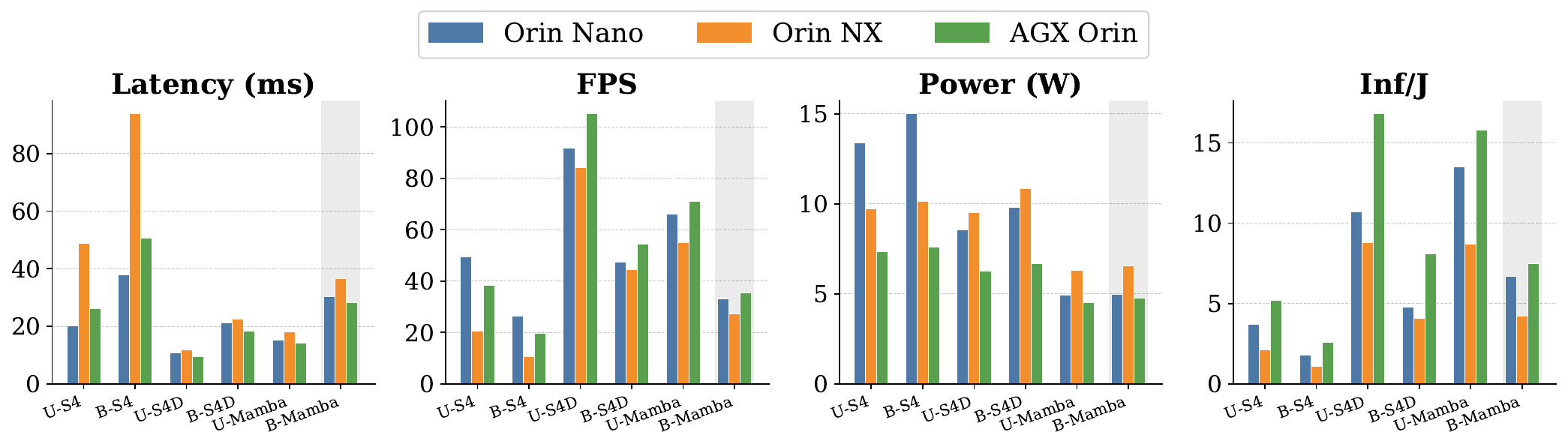}
  \caption{On-device inference comparison of SSM variants across three Jetson Orin platforms.
           Each subplot reports one metric (latency, FPS, board power, and inferences per joule)
           for all unidirectional and bidirectional S4, S4D, and Mamba models.
           Shaded group highlights Bi-Mamba (MambaGaze).
           Model labels: U- = Uni-, B- = Bi-.}
  \label{fig:benchmark_bars}
\end{figure*}

We benchmark all models on three NVIDIA Jetson Orin platforms (Nano, NX, AGX Orin) at batch size~1, sequence length~500, and 30 input channels.
Bi-Mamba achieves 27--36\,FPS across all three Jetson devices
(Nano: 33.0, NX: 27.4, AGX: 35.5), which exceeds the 0.1\,Hz sampling requirement
for real-time 10-second cognitive load windows by more than two orders of magnitude.
Latency ranges from 28\,ms (AGX) to 37\,ms (NX), with highly stable P99 values
($<$\,2\,ms above the mean), indicating stable per-inference timing.
Board power peaks at 6.56\,W on the Orin~NX, enabling battery-powered or
vehicle-embedded deployments.
Bi-S4 is the worst bidirectional model on every metric: extreme latency on the Orin~NX (93.8\,ms, 10.7\,FPS), peak power of 15.0\,W on the Nano, $3.0\times$ Bi-Mamba's 4.97\,W, and the lowest energy efficiency of any SSM at only 1.1--1.8\,Inf/J versus Bi-Mamba's 4.2--6.7\,Inf/J, consistent with unfused sequential S4 kernels that idle the GPU between steps.

Figure~\ref{fig:benchmark_bars} compares all six SSM variants across the three Jetson platforms.
Isolating the kernel effect: Bi-Mamba and Bi-S4D share the same bidirectional structure but differ only in SSM kernel (Mamba-2 vs.\ S4D). Bi-Mamba reduces board power by 29--50\% across devices (Nano: 4.97 vs.\ 9.80\,W; NX: 6.56 vs.\ 10.87\,W; AGX: 4.75 vs.\ 6.70\,W) and improves energy efficiency by 40\% on Nano (6.7 vs.\ 4.8\,Inf/J), consistent with Mamba-2's hardware-optimized CUDA scan kernels rather than architectural differences alone.
All four S4/S4D models draw 8.5--15.0\,W on the Nano versus both Mamba models' $\approx$5\,W, suggesting that kernel implementation, not directionality or GFLOP count, is the main driver of power efficiency on Jetson hardware.
Unidirectional Mamba achieves the highest energy efficiency of all models (8.7--15.8\,Inf/J), but gives up bidirectional context, which is associated with stronger cross-subject generalization in our results (Section~\ref{sec:experiments}).

\section{Conclusion}
\label{sec:conclusion}

We propose MambaGaze (Bi-Mamba), a framework for cognitive load classification from eye-tracking data with inherent missing observations. MambaGaze addresses two key challenges through XMD encoding, which augments raw features with observation masks and time-deltas to explicitly model data uncertainty, and bidirectional Mamba-2, which captures long-range temporal dependencies with linear computational complexity. Experiments on CLARE and CL-Drive under leave-one-subject-out evaluation show that MambaGaze achieves the highest average LOSO macro-F1 (55.3\%) across ten compared models, with 77.1\% accuracy and 59.2\% macro-F1 on CLARE and 69.4\% accuracy on CL-Drive. Input-stream ablation indicates that log-scaled time-deltas are the strongest single channel in our setting, and combining all three XMD streams (values, masks, deltas) provides consistent gains of 5--20\,pp macro-F1. Edge deployment benchmarks indicate real-time feasibility at 27--36\,FPS with power consumption below 6.6\,W across three NVIDIA Jetson Orin platforms. Future work will explore extending this approach to multimodal physiological signals and investigating online adaptation for personalized cognitive load thresholds.

\bibliographystyle{IEEEtran}
\bibliography{references}


\end{document}